\documentclass[10pt,twocolumn,letterpaper]{article}

\usepackage{cvpr}
\usepackage{times}
\usepackage{epsfig}
\usepackage{graphicx}
\usepackage{amsmath}
\usepackage{amssymb}

\usepackage{graphicx}
\usepackage{comment}
\usepackage{amsmath,amssymb} % define this before the line numbering.
\usepackage{color}
\usepackage{booktabs}
\usepackage{microtype}
\usepackage{epsfig}
\usepackage[tight,footnotesize]{subfigure}
\usepackage{amsthm}
\usepackage{bm}
\usepackage{multirow}
\usepackage{makecell}
\usepackage{algorithm,algpseudocode}

\usepackage{comment}
\usepackage[pagebackref=true,breaklinks=true,letterpaper=true,colorlinks,bookmarks=false]{hyperref}

\cvprfinalcopy % *** Uncomment this line for the final submission

\def\cvprPaperID{1676} % *** Enter the CVPR Paper ID here

\ifcvprfinal\fi
\begin{document}

%%%%%%%%% TITLE
\title{Revisiting Batch Normalization for Training Low-latency \\ Deep Spiking Neural Networks from Scratch}

\author{Youngeun Kim\\
Yale University\\
{\tt\small youngeun.kim@yale.edu}
\and
Priyadarshini Panda\\
Yale University\\
{\tt\small priya.panda@yale.edu}
}

\maketitle
%\thispagestyle{empty}

%%%%%%%%% ABSTRACT
\begin{abstract}
\vspace{-3mm}
Spiking Neural Networks (SNNs) have recently emerged as an alternative to deep learning owing to sparse, asynchronous and binary event (or spike) driven processing, that can yield huge energy efficiency benefits on neuromorphic hardware.
%
% Most existing approaches to create SNNs either convert the weights from pre-trained Artificial Neural Networks (ANNs) or directly train SNNs with surrogate gradient backpropagation. 
% Each approach presents its pros and cons.
% The ANN-to-SNN conversion method requires at least hundreds of time-steps for inference to yield competitive accuracy that in turn reduces the energy savings. 
% Training SNNs with surrogate gradients from scratch reduces the latency or total number of time-steps, but the training becomes slow/problematic and has convergence issues.
% Thus, the latter approach of training SNNs has been limited to shallow networks on simple datasets.
However, training high-accuracy and low-latency SNNs from scratch  suffers from non-differentiable nature of a spiking neuron. 
To address this training issue in SNNs, we revisit batch normalization and propose a  temporal Batch Normalization Through Time (BNTT) technique. 
% to reduce the latency and enable direct training deep SNNs from scratch on more complicated datasets.
% , which had not been used for SNNs so far.
%
% By modulating the statistics in all Batch Normalization layers across the network, our approach achieves deep adaptation effect for domain adaptation tasks.
%
Most prior SNN works till now have disregarded batch normalization deeming it ineffective for training temporal SNNs. Different from previous works, our proposed BNTT decouples the parameters in a BNTT layer along the time axis to capture the temporal dynamics of spikes.
%
% Training with BNTT brings several advantages:
% 1) The BNTT layer reduces internal covariate shift so that the networks  can be trained stablly from scratch even for large-scale datasets.
% 2)  Learnable parameters in BNTT enables the networks to be implemented with low latency.
% Since the networks can be implemented with low latency, our method small energy consumption compared with  .
% 3) BNTT captures the temporal dynamics of the time-sequential spikes.
%
The temporally evolving learnable parameters in BNTT allow a neuron to control its spike rate through different time-steps, enabling low-latency and low-energy training from scratch.
% to reduce the latency and enable direct training deep SNNs from scratch on more complicated datasets.
% train with low-latency and low-energy consumption.
%
% Therefore,  based on the distribution of a parameter in BNTT, we propose an early exit algorithm to further reduce the latency at inference.
%
We conduct experiments on CIFAR-10, CIFAR-100, Tiny-ImageNet and event-driven DVS-CIFAR10 datasets. 
% Our method reduces the latency (more than 4$\times$ faster compared to state-of-the-art SNNs) and energy consumption (9$\times$ less compared to  standard  ANN) significantly, while preserving competitive classification accuracy. 
BNTT allows us to train deep SNN architectures from scratch, for the first time, on complex datasets with just few 25-30 time-steps.
We also propose an early exit algorithm using the  distribution of parameters in BNTT to reduce the latency at inference, that further improves the energy-efficiency. 
The code has been released at https://github.com/Intelligent-Computing-Lab-Yale/BNTT-Batch-Normalization-Through-Time.
% \textit{ Code will be made available}.
\end{abstract}
\vspace{-3mm}
%%%%%%%%% BODY TEXT

\section{Introduction}
% \vspace{-3m}
%P1: Beauty of SNN

Artificial Neural Networks (ANNs) have shown  state-of-the-art performance across various computer vision tasks. Nonetheless, huge energy consumption incurred for implementing ANNs on conventional von-Neumann hardware limits their usage in low-power and resource-constrained Internet of Things (IoT) environment, such as mobile phones, drones among others.
In the context of low-power machine intelligence, Spiking Neural Networks (SNNs)  have received considerable attention in the recent past \cite{roy2019towards,panda2020toward,cao2015spiking,diehl2015unsupervised,comsa2020temporal}.
Inspired by biological neuronal mechanisms, SNNs process visual information with discrete spikes or events over multiple time-steps. Recent works have shown that the event-driven behavior of SNNs can be implemented on emerging neuromorphic hardware to yield 1-2 order of magnitude energy efficiency over ANNs \cite{akopyan2015truenorth,davies2018loihi}.
%This event-driven behavior can be efficiently implemented on  to emerging neuromorphic hardware (\cite{akopyan2015truenorth,davies2018loihi}).
% The Leaky Integrate and Fire (LIF) neuron is widely used for modeling SNNs, which consists of the membrane potential with leaky behavior.
% The membrane accumulates the input spikes over time when the potential is below the threshold value, otherwise, generates an output binary spike.
Despite the energy efficiency benefits, SNNs have still not been widely adopted due to inherent training challenges.
% Training SNNs to yield competitive accuracy as ANNs remains a huge challenge today.
The training issue arises from the non-differentiable characteristic of a spiking neuron, generally, Integrate-and-Fire (IF) type \cite{burkitt2006review}, that makes SNNs incompatible with gradient descent training.
%problematic due to its non-differentiable characteristic of a spiking neuron.

%----------------------------------------
\begin{figure*}[t]
\begin{center}
\def\arraystretch{0}
\begin{flushright}
\begin{tiny}
{*Method (total time-steps / accuracy)}
\end{tiny}
\end{flushright}
\vspace*{-0.03in}
\begin{tabular}{@{\hskip 0.05\linewidth}c@{\hskip 0.05\linewidth}c@{\hskip 0.05\linewidth}c}
\includegraphics[width=0.26\linewidth]{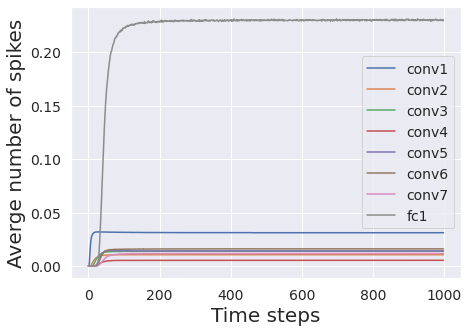} &
\includegraphics[width=0.26\linewidth]{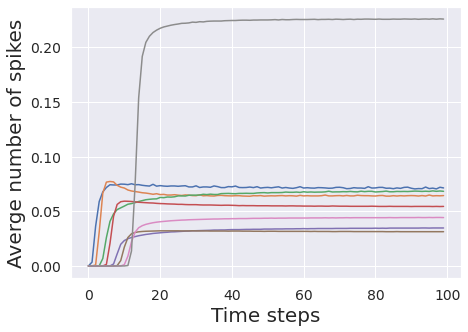}  &
\includegraphics[width=0.26\linewidth]{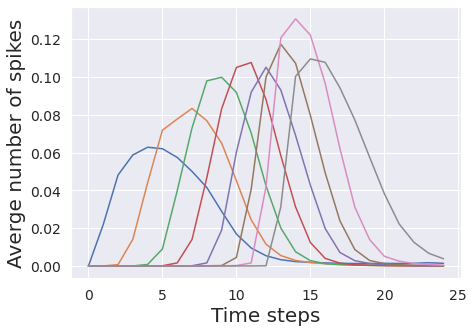}
\\
{\hspace{4.3mm} (a) Conversion (1000 / 91.2\%)} & {\hspace{4.5mm}(b) Surrogate BP (100 / 88.7\%) } & {\hspace{4.5mm}(c) BNTT (25 / 90.5\%) }\\
\end{tabular}
% \vspace*{-0.1in}
\end{center}
\caption
{
Visualization of the average number of spikes in each layer with respect to time-steps. 
% We also report  (total time-steps /  accuracy) for each method.
Compared to (a) ANN-SNN conversion  and (b) surrogate gradient-based backpropagation, our (c) BNTT captures the temporal dynamics  of spike activation with learnable parameters, enabling low-latency (\ie, small time-steps) and low-energy (\ie, less number of spikes) training.
All experiments are conducted on CIFAR-10 with VGG9.
}
% \vspace{-4mm}
\label{fig:motivation}
\end{figure*}
%----------------------------------------

%P2:  Previous approaches

To address the training issue of SNNs, several methods, such as, \textit{Conversion} and \textit{Surrogate Gradient Descent} have been proposed. In ANN-SNN conversion \cite{sengupta2019going,han2020rmp,diehl2015fast,rueckauer2017conversion}, off-the-shelf trained ANNs are converted to SNNs using normalization methods to transfer ReLU activation to IF spiking activity. The advantage here is that training happens in the ANN domain leveraging widely used machine learning frameworks like, PyTorch, that yield short training time and can be applied to complex datasets.  
%Converting SNNs from off-the-shelf trained ANNs can leverage well-organized  ANN training methods with short training time. 
But the ANN-SNN conversion method  requires large number of time-steps ($\sim 500-1000$) for inference to yield competitive accuracy, which significantly increases the latency and energy consumption of the SNN.
On the other hand, directly training SNNs with a surrogate gradient function \cite{neftci2019surrogate,lee2020enabling,wu2018spatio} exploits temporal dynamics of spikes, resulting in  lesser number of time-steps ($\sim 100-150$).
However, the discrepancy between forward spike activation function and backward surrogate gradient function during backpropagation restricts the training capability. Only shallow SNNs (\eg, VGG5) can be trained using surrogate gradient descent and therefore they achieve high performance only for simple datasets (\eg, MNIST and CIFAR-10).
Recently, a hybrid method \cite{rathi2020enabling} that combines the conversion method and the surrogate gradient-based method shows state-of-the-art performance at reasonable latency ($\sim 250$ time-steps). However, the hybrid method incurs sequential processes, \ie, training ANN from scratch, conversion of ANN to SNN, and training SNNs using surrogate gradient descent, that increases the total computation cost to obtain the final SNN model.
Overall, training high-accuracy and low-latency SNNs from scratch still remains an open problem.

%P3:  BN for SNN
In this paper, we revisit Batch Normalization (BN) for more advanced SNN training.
The BN layer \cite{ioffe2015batch} has been used extensively in deep learning to accelerate the training process of ANNs.
It is well known that BN reduces internal covariate shift (or soothing optimization landscape \cite{santurkar2018does}) mitigating the problem of exploding/vanishing gradients.
However, till now, numerous studies on surrogate gradient of SNNs \cite{lee2016training} have witnessed that BN does not help with SNN optimization. %\textcolor{red}{Therefore, such works rely on a dropout technique while training.}
Moreover, most ANN-SNN conversion methods \cite{sengupta2019going} get rid of BN since  time-sequential spikes with BN set the firing threshold of all neurons to non-discriminative/similar values across all inputs, resulting in accuracy decline.
%Also, it is known for the BN layer does not address the sparse spike problem in deeper layers, which is the chronic problem of surrogate gradient-based methods.

%P4:  Our method
\textbf{Motivation \& Contribution:} A natural question then arises: \textit{Can standard BN capture the proper structure of temporal dynamics of spikes in SNNs?}
Through this paper, we assert that standard BN hardly captures temporal characteristics as it represents the statistics of total time-steps as one common parameter. Thus, a  {temporally adaptive} BN approach  is  required.
% Our main hypothesis on this issue is that  spike activation of each time step has totally different characteristic.
% For example, spike activation at layer $5$ is zero before time-step $5$ until spikes arrive and then the activation gradually increases.
% Therefore, standard BN hardly captures temporal characteristics as it represents the statistics of total time-steps as one common parameter,  so  a  new  approach  is  required.
To this end, we propose a new SNN-crafted batch normalization layer called Batch Normalization Through Time (BNTT) that decouples the parameters in the BN layer  across different time-steps.
BNTT is implemented as an additional layer in SNNs and is trained with surrogate gradient backpropagation.
%
% This temporally uncorrelated learnable parameters enable the networks to deal spike information effectively.
%
To investigate the effect of our BNTT, we compare the statistics of spike activity of BNTT with previous approaches: Conversion \cite{sengupta2019going} and standard Surrogate Gradient Descent \cite{neftci2019surrogate}, as shown in Fig. \ref{fig:motivation}.
Interestingly, different from the conversion method and surrogate gradient method (without BNTT) that maintain reasonable spike activity  during the entire time period across different layers,  spike activity of layers trained with BNTT follows a gaussian-like trend. BNTT imposes a variation in spiking across different layers, wherein, each layer's activity peaks in a particular time-step range and then decreases.
Moreover, the peaks for early layers occur at initial time-steps and latter layers peak at later time-steps. 
This phenomenon implies that learnable parameters in BNTT enable the networks to pass the visual information temporally from shallow  to deeper layers in an effective manner.

%P5:  Advantages
% Combining the newly observed characteristics of our BNTT with standard BN brings several advantages.
The newly observed characteristics of BNTT brings several advantages.
First, similar to BN, the BNTT layer enables 
% reduces internal covariate shift 
 SNNs to be trained stably from scratch even for large-scale datasets.
Second, learnable parameters in BNTT enable SNNs to be trained with low latency ($\sim 25-50$ time-steps) and impose optimum spike activity across different layers for low-energy inference.
Finally, the distribution of the BNTT learnable parameter  (\ie, $\gamma$) is a good representation of the temporal dynamics of spikes. Hence, relying on the observation that low $\gamma$ value induces low spike activity and vice-versa, we further propose a temporal early exit algorithm. Here, an SNN can predict at an earlier time-step and does not need to wait till the end of the time period to make a prediction. %all $\gamma$ values  are lower than the predefined threshold value.

% Therefore,  based on the distribution of a learnable parameter (\ie, gamma) in BNTT, we further propose an early exit algorithm to further reduce the latency at inference.

%P6:  Contributions
In summary, our key contributions are as follows: 
(i) For the first time, we introduce a batch normalization technique for SNNs, called BNTT.
(ii) BNTT allows SNNs to be implemented in a low-latency and low-energy environment. 
(iii) We further propose a temporal early exit algorithm at inference time by monitoring the learnable parameters in BNTT.
(iv) To ascertain that BNTT captures the temporal characteristics of SNNs, we mathematically show that proposed BNTT has similar effect as controlling the firing threshold of the spiking neuron at every time step during inference.

\vspace{-2mm}

% \newpage
\section{Batch Normalization}
\vspace{-2mm}

Batch Normalization (BN) reduces the internal covariate shift (or variation of loss landscape \cite{santurkar2018does}) caused by the distribution change of input signal, which is a known problem of deep neural networks \cite{ioffe2015batch}.
Instead of calculating the statistics of total dataset, the intermediate representations are standardized with a mini-batch to reduce the computation complexity.
Given a mini-batch $\mathcal{B} = \{x_{1,...,m}\}$, the BN layer computes the mean and variance of the mini-batch as:
\begin{equation}
    \mu_{\mathcal{B}} = \frac{1}{m} \sum_{b=1}^{m} x_b ;
    \hspace{10mm}
    \sigma_{\mathcal{B}}^2 = \frac{1}{m} \sum_{b=1}^{m} (x_b - \mu_\mathcal{B})^2.
    \label{eq:BN_mu}
\end{equation}
% \begin{equation}
%     \sigma_{\mathcal{B}}^2 = \frac{1}{m} \sum_{i=1}^{m} (x_i - \mu_\mathcal{B})^2.
%     \label{eq:BN_sigman}
% \end{equation}
Then, the input features in the mini-batch are normalized with calculated statistics as:
\begin{equation}
    {\widehat{x}_b} = \frac{x_b - \mu_\mathcal{B} }{\sqrt{\sigma_{\mathcal{B}}^2 + \epsilon}},
    \label{eq:BN_normalize}
\end{equation}
where, $\epsilon$ is a small constant for numerical stability.
To further improve the representation capability of the layer, learnable parameters $\gamma$ and $\beta$ are used to transform the input features that can be formulated as $BN(x_i) = \gamma {\widehat{x}_i}+ \beta$.
% \begin{equation}
    % BN(x_i) = \gamma {\widehat{x}_i}+ \beta.
    % \label{eq:BN_remapping}
% \end{equation}
At inference time, BN uses the running average of mean and variance obtained from training.
Previous works show that the BN layer not only improves the performance but also reduces the number of iterations required for training convergence.
Therefore, BN is an indispensable  training component for all ANN models, such as convolutional neural networks \cite{simonyan2014very} and recurrent neural networks \cite{greff2016lstm}.
On the other hand, the effectiveness of BN in  bio-plausible SNNs has not been observed yet.

% \begin{wrapfigure}{o}{0.2\linewidth}
%   \begin{center}
%     \includegraphics[width=1.0\linewidth]{figures/AGD.png}
%   \end{center}
%   \caption{Birds}
% \end{wrapfigure}

% \begin{wrapfigure}[r]{0.2\textwidth}
%     % \begin{center}
%     % \def\arraystretch{0}
%     % \vspace*{-0.03in}
%     % \begin{tabular}{@{}c@{\hskip 0.07\linewidth}c@{}c}
%     % \includegraphics[width=0.55\linewidth]{figures/spike_mach.png} &
%     % \includegraphics[width=0.35\linewidth]{figures/AGD.png} 
%     % \\
%     % {\hspace{4.3mm} (a) } & {\hspace{4.5mm}(b)  } \\
%       \includegraphics[width=0.35\textwidth]{figures/AGD.png}
%     % \end{tabular}
%     % \vspace*{-0.1in}
%     % \end{center}
%     \caption
%     {
%     (a) \citetivities in Leaky-Integrate-and-Fire neurons. 
%     (b) The approximated gradient value with respect to the membrane potential.
%     }
%     % \vspace{-3mm}
%     % \label{fig:spike_optimization}
% \end{wrapfigure}

\section{Methodology}

\vspace{-0mm}

\subsection{Spiking Neural Networks}

\vspace{-0mm}

Different from conventional ANNs, SNNs transmit information using binary spike trains.
To leverage the temporal spike information,  Leaky-Integrate-and-Fire (LIF) model \cite{dayan2001theoretical} is widely used to emulate neuronal functionality in SNNs, which can be formulated as a differential equation:
\begin{equation}
    \tau_m \frac{dU_m}{dt} = -U_m  + RI(t),
    \label{eq:LIF_origin}
\end{equation}
where, $U_m$ represents  the  membrane  potential  of  the  neuron that characterizes the internal state of the neuron, $\tau_m$ is the time constant of membrane potential decay.
Also, $R$ and $I(t)$ denote the input resistance and the input current at time $t$,  respectively. 
Following the previous work \cite{wu2019direct}, we convert this continuous dynamic equation into a discrete equation for digital simulation.
For a single post-synaptic neuron $i$, we can represent the membrane potential $u_i^t$ at time-step $t$ as:
\begin{equation}
    u_i^t = \lambda u_i^{t-1} + \sum_j w_{ij}o^t_j.
    \label{eq:LIF}
\end{equation}
Here, $j$ is the index of a pre-synaptic neuron, $\lambda$ is a leak factor with value less than $1$,
$o_j$ is the binary spike activation, and $w_{ij}$ is the weight of the connection between pre- and post-neurons.
From Eq. \ref{eq:LIF}, the membrane potential of a neuron decreases due to leak and increases due to the weighted sum of incoming input spikes.

\begin{figure}[t]
  \begin{center}
    \includegraphics[width=0.5\textwidth]{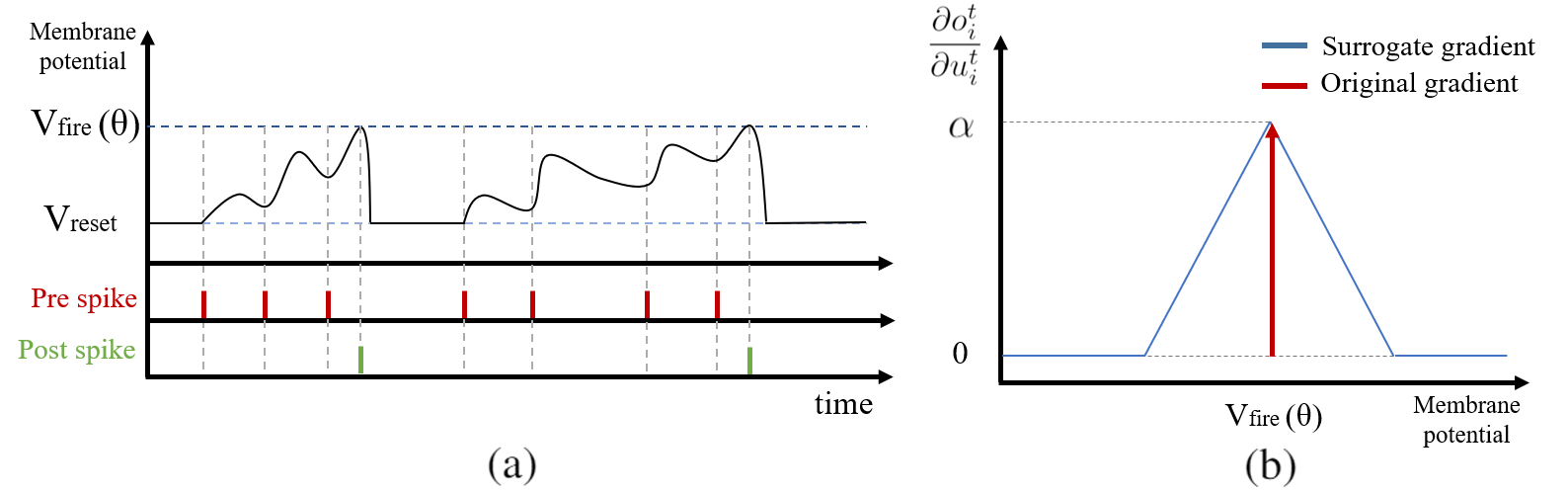}
  \end{center}
     \vspace{-2mm}
  \caption{ (a) Illustration of spike activities in Leaky-Integrate-and-Fire neurons. 
     (b) The approximated gradient value with respect to the membrane potential.}
   \label{fig:spike_optimization}
%   \vspace{-3mm}
\end{figure}

If the membrane potential $u$ exceeds a pre-defined firing threshold $\theta$, the LIF neuron $i$ generates a binary spike output $o_i$. 
After that, we perform a soft reset, where the membrane potential $u_i$ is reset by reducing its value by the threshold $\theta$.
Compared to a hard reset (resetting the membrane potential $u_i$ to zero after neuron $i$ spikes), the soft reset minimizes information loss by maintaining the residual voltage and carrying it forward to the next time step, thereby achieving better performance \cite{han2020rmp}. 
Fig. \ref{fig:spike_optimization}(a) illustrates the membrane potential dynamics of a LIF neuron.

For the output layer, we discard the thresholding functionality so that neurons do not generate any spikes. We allow the output neurons to accumulate the spikes over all time-steps by fixing the leak parameter ($\lambda$ in Eq. \ref{eq:LIF}) as one. This enables the output layer to compute probability distribution after softmax function without information loss.
As with ANNs, the number of output neurons in SNNs is identical to the number of classes $C$ in the dataset.
From the accumulated membrane potential, we can define the cross-entropy loss for SNNs as:
\begin{equation}
    {L} = - \sum_{i} y_{i} log(\frac{e^{u_i^T}}{\sum_{k=1}^{C}e^{u_k^T}}),
    \label{eq:celoss}
\end{equation}
where, $y$ is the ground-truth label, and $T$ represents the total number of time-steps. Then, the weights of all layers are updated 
by backpropagating the loss value with gradient descent.
% measured by a difference between labels and probabilities, we update the weights of all layers

%Backpropagation
To compute the gradients of each layer $l$, we use back-propagation through time (BPTT), which accumulates the gradients over all time-steps \cite{neftci2019surrogate}.
These approaches can be implemented with auto-differentiation tools, such as PyTorch \cite{paszke2017automatic}, that enable  backpropagation on the unrolled network.
To this end, we compute the loss function at time-step $T$ and use gradient descent optimization.
Mathematically, we can define the accumulated gradients at the  layer $l$ by chain rule as:
% \begin{equation}
%     \Delta W_l = \sum_{t} \frac{\partial L}{\partial W_l^t} =
% \begin{cases}
%  \sum_{t}\frac{\partial L}{\partial o_i^t}\frac{\partial o_i^t}{\partial u_i^t}\frac{\partial u_i^t}{\partial W_l^t},              & \text{if $l$ = hidden layer} \\
%     \sum_{t}\frac{\partial L}{\partial u_i^T}\frac{\partial u_i^T}{\partial W_l^t}.
%     & \text{if $l$ = output layer} 
% \end{cases}
% \label{eq:delta_W}
% \end{equation}
\begin{equation}
    \Delta W_l = \sum_{t} \frac{\partial L}{\partial W_l^t} =
\begin{cases}
 \sum_{t}\frac{\partial L}{\partial O_l^t}\frac{\partial O_l^t}{\partial U_l^t}\frac{\partial U_l^t}{\partial W_l^t},              & \text{if $l$ = hidden layer} \\
    \sum_{t}\frac{\partial L}{\partial U_l^T}\frac{\partial U_l^T}{\partial W_l^t}.
    & \text{if $l$ = output layer} 
\end{cases}
\label{eq:delta_W}
\end{equation}
Here, $O_l$ and $U_l$ are output spikes  and membrane potential at layer $l$, respectively.
For the output layer, we get the derivative of the loss $L$ with respect to the membrane potential $u_i^T$ at final time-step $T$:
\begin{equation}
\frac{\partial L}{\partial u_i^T} =\frac{e^{u_i^T}}{\sum_{k=1}^{C}e^{u_k^T}} - y_i.
\end{equation}
This derivative function is continuous and differentiable for all possible membrane potential values.
On the other hand, LIF neurons in hidden layers  generate  spike output only if the membrane potential $u_i^t$ exceeds the firing threshold, leading to non-differentiability.
To deal with this problem, we introduce an approximate gradient:
\begin{equation}
    \frac{\partial o_i^t}{\partial u_i^t} = \alpha \max \{0, 1-  \ | \frac{u_i^t - \theta}{\theta} \ | \},
\end{equation}
where, $\alpha$ is a damping factor for back-propagated gradients.
% , as shown in Fig. \ref{fig:spike_optimization}(b).
Note, a large $\alpha$ value causes unstable training as gradients are summed over all time-steps. Hence, we set $\alpha$ to $0.3$.
Overall, we update the network parameters at the layer $l$ based on the gradient value (Eq. \ref{eq:delta_W}) as $W_l = W_l - \eta \Delta W_l$.

% \newpage
\subsection{Batch Normalization Through Time (BNTT)}

The main contribution of this paper is a new SNN-crafted Batch Normalization (BN) technique.
Naively applying BN does not have any effect on training SNNs. This is because using the same BN parameters (\eg, global mean $\mu$, global variation $\sigma$, and learnable parameter $\gamma$) for the statistics of all time-steps  do not capture the temporal dynamics of input spike trains. 
% A standard BN layer did not show its effectiveness for training SNNs as each LIF neuron receives signals from  various distributions across time, hence, naively applying a BN layer in SNNs may lose its temporal dynamics.
For example, an LIF neuron requires at least one time-step to propagate spikes to the next layer. Therefore, input signals for the third layer of an SNN will have a zero value till $t=2$. Following the initial spike activity in the layer at $t=2$, the spike signals vary depending upon the weight connections and the membrane potentials of previous layers.
Therefore, a fixed global mean from a standard BN layer may not store any time-specific information, resulting in performance degradation at inference.

To resolve this issue, we vary the internal parameters in a BN layer through time, that we define as, BNTT.
Similar to the digital simulation of LIF neuron across different time-steps,
% global mean, and global variance, and learnable parameter $\gamma$ (we fix $\beta$ to zero) are expanded  
one BNTT layer is expanded temporally with a local learning parameter associated with each time-step. This allows the BNTT layer to capture temporal statistics (see Section 3.3 for mathematical analysis).
%This temporal independent statistic allows the networks to capture the temporal information.
The proposed BNTT layer is easily applied to SNNs by inserting the layer after convolutional/linear operations as:
\begin{equation}
    \begin{split}
    u_i^t = & \lambda u_i^{t-1} + BNTT_{\gamma^t}(\sum_j w_{ij}o^t_j ) \\ 
    = & \lambda u_i^{t-1} + \gamma_i^t (\frac{\sum_j w_{ij}o^t_j  - \mu^t_i}{\sqrt{(\sigma_i^t)^2 + \epsilon}}).
    \end{split}
    \label{eq:LIFwithBN}
\end{equation}

During the training process,
we compute the mean $\mu_{i}^t$ and variance $\sigma_{i}^t$ from the samples in a mini-batch $\mathcal{B}$ for each time step $t$, as shown in Algorithm 1.
Note, for each time-step $t$, we apply an exponential moving average to approximate global mean $\Bar{\mu}_{i}^t$ and variance $\Bar{\sigma}_{i}^t$ over training iterations. These global statistics are used to normalize the test data at inference. Also, we do not utilize $\beta$ as in conventional BN, since it adds redundant voltage to the membrane potential of SNNs.

Adding the BNTT layer to LIF neurons changes the gradient calculation for backpropagation.
Given that $x_i^t = \sum_j w_{ij}o^t_j  $ is an input signal to the BNTT layer, we can calculate the gradient value  passed through lower layers by the BNTT layer as:
% Given that $x_i^t = \sum_j w_{ij}o^t_j  $ is an input signal to the BNTT layer, we can reformulate the accumulated gradients at layer $l$ (Eq. \ref{eq:delta_W}) by multiplying the derivative of the membrane potential with respect to the weighted summed input signal:
% \begin{equation}
% \vspace{-1mm}
%     \Delta W_l =  \sum_{t}\frac{\partial L}{\partial O_l^t}\frac{\partial O_l^t}{\partial U_l^t}\frac{\partial  U_l^t}{\partial X_l^t}\frac{\partial X_l^t}{\partial W_l^t},
%     % \vspace{-1mm}
% \end{equation}
% where,  $\frac{\partial  U_l^t}{\partial  X_l^t}$ is the backward gradient of the BNTT layer (see Appendix A for more detail), which can be formulated as:
\begin{equation}
\vspace{0mm}
\frac{\partial  L}{\partial  x_b^t}
 = \frac{1}{m \sqrt{(\sigma^t)^2 + \epsilon}} 
\left(
m \frac{\partial  L}{\partial  \hat{x}_b^t} 
-
\sum_{k=1}^{m} \frac{\partial  L}{\partial  \hat{x}_k^t}
-
\hat{x}^t_b \sum_{k=1}^{m} \frac{\partial  L}{\partial  \hat{x}_k^t}\hat{x}_k^t
\right).
% \vspace{-1mm}
\end{equation}
Here, we omit a neuron index $i$ for simplicity. Also, $m$ and $b$ denote the batch size and batch index (see Appendix A for more detail).
Thus, for every time-step $t$, gradients are calculated based on the  time-specific statistics of input signals.
This allows the networks to take into account temporal dynamics for training weight connections.
Moreover, a learnable parameter $\gamma$ is updated to restore the representation power of the batch normalized signal. 
Since we use different $\gamma^t$ values across all time-steps,
$\gamma^t$ finds an optimum over each time-step for efficient inference.  We update gamma $\gamma^t  = \gamma^t  - \eta \Delta \gamma^t$ where:
% The update process of $\gamma^t$ can be formulated as:
\begin{equation}
\Delta \gamma^t =  \frac{\partial  L}{\partial  \gamma^t} =  \frac{\partial  L}{\partial  u^t} \frac{\partial  u^t}{\partial  \gamma^t} = \sum_{k=1}^{m} \frac{\partial  L}{\partial  u_k^t} \hat{x}^t_k.
% = 
% \sum_{j=1}^{m} \frac{\partial  L}{\partial  u_j^t} \frac{x^t_j -\mu^t }{\sqrt{(\sigma^t)^2 + \epsilon}}.
\end{equation}
% \begin{equation}
% \gamma^t  = \gamma^t  - \eta \Delta \gamma^t.
% \end{equation}
% Again, the derivation of gamma depends on only time step $t$, varying signal 
% Algorithm 2 summarizes the whole training process of SNNs with our BNTT.

\subsection{Mathematical Analysis}
\label{sec: mathmaticalanalysis}
In this section, we discuss the connections between BNTT and the firing threshold of a LIF neuron. Specifically, we formally prove that using BNTT has a similar effect as varying the firing threshold over different time-steps, thereby ascertaining that BNTT captures temporal characteristics in SNNs.
% Recall that a neuron produces a output spike activation whenever the membrane potential exceeds the pre-defined firing threshold $\theta$. 
Recall that BNTT normalizes the input signal using 
stored approximated global average $\bar{\mu}_i^t$ and standard deviation $(\bar{\sigma_i}^t)^2$ at inference.
From Eq. \ref{eq:LIFwithBN}, we can calculate a membrane potential at time-step $t=1$, given that initial membrane potential $u_i^0$ has a zero value:
\begin{equation}
    \begin{split}
        u_i^1 = & \gamma_i^1 (\frac{\sum_j w_{ij}o^1_j  - \bar{\mu}^1_i}{\sqrt{(\bar{\sigma}_i^1)^2 + \epsilon}}) \\
    \approx  & \frac{\gamma_i^1}{\sqrt{(\bar{\sigma}_i^1)^2 + \epsilon}} {\sum_j w_{ij}o^1_j}
    = \frac{\gamma_i^1}{\sqrt{(\bar{\sigma}_i^1)^2 + \epsilon}} \tilde{u}_i^{1}.
    \end{split}
    \label{eq: prove1}
\end{equation}
Here, we assume $\bar{\mu}^1_i$ can be neglected with small signal approximation due to the spike sparsity in SNNs, and $\tilde{u}_i^{1}=\sum_j w_{ij}o^1_j$ is membrane potential at time-step $t=1$ without BNTT (obtained from Eq. \ref{eq:LIF}). We can observe that 
the membrane potential with BNTT is proportional to the membrane potential without BNTT  at $t=1$.
For time-step $t > 1$, we should take into account the membrane potential from the previous time-step, which is multiplied by leak $\lambda$.
To this end, by substituting Eq. \ref{eq: prove1} in the BNTT equation (Eq. \ref{eq:LIFwithBN}), 
we can formulate the membrane potential at $t=2$ as:
\begin{equation}
    \begin{split}
    u_i^2 & \approx  \lambda  u_i^1 + \frac{\gamma_i^2}{\sqrt{(\sigma_i^2)^2 + \epsilon}} {\sum_j w_{ij}o^2_j}  \\
    & =      (\frac{\lambda  \gamma_i^1}{\sqrt{(\sigma_i^1)^2 + \epsilon}}) \tilde{u}_i^1 + \frac{\gamma_i^2}{\sqrt{(\sigma_i^2)^2 + \epsilon}} {\sum_j w_{ij}o^2_j} \\
    & \approx     \frac{\gamma_i^2}{\sqrt{(\sigma_i^2)^2 + \epsilon}} \{\lambda   \tilde{u}_i^1 +  {\sum_j w_{ij}o^2_j} \} = \frac{\gamma_i^2}{\sqrt{(\sigma_i^2)^2 + \epsilon}}  \tilde{u}_i^2.
    \end{split}
\end{equation}

\begin{algorithm}[t]\small
    \caption{BNTT layer}
%   \hspace*{\algorithmicindent} 
   \textbf{Input}: mini-batch $\mathcal{B}$ at time step $t$ (${x^{t}_{\{1...m\}}}$),
   learnable parameter ($\gamma^t$), update factor ($\alpha$)               
   \\
  \textbf{Output}:  $\{y^t = \textup{BNTT}_{\gamma^t}(x^t)\}$
  \begin{algorithmic}[1]
    % \State{\textbf{begin}}
    %
    \vspace{1mm}
\State{$\mu^t \leftarrow \frac{1}{m} \sum^{m}_{b=1}x^t_{b}$}
    \vspace{1mm}
    \State{$(\sigma^t)^2 \leftarrow \frac{1}{m} \sum^{m}_{b=1}(x_{b}^t-\mu^t)^2$}
    \vspace{1mm}
    \State{$\hat{x}^t = \frac{x^t -\mu^t }{\sqrt{(\sigma^t)^2 + \epsilon}}$}
    \vspace{1mm}
    \State{$y^t \leftarrow \gamma^t \hat{x}^t  \equiv \textup{BNTT}_{\gamma^t}(x^t)$}
    \vspace{1.5mm}
    \State{\% Exponential moving average}
    \vspace{1mm}
    \State{$\Bar{\mu}^t \leftarrow (1-\alpha)\Bar{\mu}^t + \alpha {\mu}^t$ }
    \vspace{1mm}
     \State{$\Bar{\sigma}^t \leftarrow (1-\alpha)\Bar{\sigma}^t + \alpha {\sigma}^t$ }
  \end{algorithmic}
      \label{algorithm: BNTT}
\end{algorithm}

In the third line, the learnable parameter $\gamma^t_i$ and $\sigma^t_i$ have similar values in adjacent time intervals ($t= 1, 2$) because of continuous time property. Hence, we can approximate $\gamma^1_i$ and $\sigma^1_i$  as $\gamma^2_i$ and $\sigma^2_i$, respectively.
Finally, we can extend the equation of BNTT  to the time-step $t$:
\begin{equation}
   {u}_i^t  \approx \frac{\gamma_i^t}{\sqrt{(\sigma_i^t)^2 + \epsilon}}  \tilde{u}_i^t.
   \label{eq: prove_memprop}
\end{equation}
Considering that a neuron produces an output spike activation whenever the membrane potential $\tilde{u}^{t}_i$ exceeds the pre-defined firing threshold $\theta$,
the spike firing condition with BNTT can be represented ${u}^{t}_i \geq {\theta}$.
Comparing with the threshold of a neuron without BNTT,  we can reformulate the firing condition as:
\begin{equation}
   \tilde{u}^{t}_i \geq \frac{\sqrt{(\sigma_i^{t})^2 + \epsilon}}{\gamma_i^{t}}
{\theta}.
    \label{eq: firing_condition}
\end{equation}
Thus, we can infer that using a BNTT layer changes the  firing threshold value by $ \sqrt{(\sigma_i^t)^2 + \epsilon}/{\gamma_i^t}$ at every time-step. 
%In other words,
In practice, BNTT results in an optimum $\gamma$ during training that improves the representation power, producing better performance and low-latency SNNs.%, similar to that of changing the firing threshold of each neuron of the SNN at each time-step.
This observation allows us to consider the advantages of time-varying learnable parameters in SNNs.
This implication is in line with previous work \cite{han2020rmp}, which insists that manipulating the firing threshold improves the performance and latency of the ANN-SNN conversion method. However, Han et al. change the threshold value in a heuristic way without any optimization process and fix the threshold value across all time-steps. On the other hand, our BNTT yields time-specific $\gamma^{t}$ which can be optimized via back-propagation.

% \hspace{-4mm}
% \begin{minipage}[!t]{.5\textwidth}
%     \vspace{-0.6cm}
%     \begin{algorithm}[H]\small
%         \caption{BNTT layer}
%     %   \hspace*{\algorithmicindent} 
%       \textbf{Input}: mini-batch $\mathcal{B}$ at time step $t$ (${x^{t}_{\{1...m\}}}$),
%       learnable parameter ($\gamma^t$), update factor ($\alpha$)`               
%       \\
%       \textbf{Output}:  $\{y^t = \textup{BNTT}_{\gamma^t}(x^t)\}$
%       \begin{algorithmic}[1]
%         % \State{\textbf{begin}}
%         %
%         \vspace{1mm}
%     \State{$\mu^t \leftarrow \frac{1}{m} \sum^{m}_{b=1}x^t_{b}$}
%         %
%         \vspace{1mm}
%         \State{$(\sigma^t)^2 \leftarrow \frac{1}{m} \sum^{m}_{b=1}(x_{b}^t-\mu^t)^2$}
%         %
%         \vspace{1mm}
%         \State{$\hat{x}^t = \frac{x^t -\mu^t }{\sqrt{(\sigma^t)^2 + \epsilon}}$}
%         \vspace{1mm}
%         \State{$y^t \leftarrow \gamma^t \hat{x}^t  \equiv \textup{BNTT}_{\gamma^t}(x^t)$}
%         \vspace{1.5mm}
%         \State{\% Exponential moving average}
%         \vspace{1mm}
%         \State{$\Bar{\mu}^t \leftarrow (1-\alpha)\Bar{\mu}^t + \alpha {\mu}^t$ }
%         \vspace{1mm}
%          \State{$\Bar{\sigma}^t \leftarrow (1-\alpha)\Bar{\sigma}^t + \alpha {\sigma}^t$ }
%       \end{algorithmic}
%           \label{algorithm: BNTT}
%     \end{algorithm}
% \end{minipage}
%
% \vspace{-1m}
%  \hspace{4mm}
% \begin{minipage}[!t]{.5\textwidth}
    % \vspace{-0.3cm}
\begin{algorithm}[t]\small
    \caption{Training process with BNTT }
%   \hspace*{\algorithmicindent} 
   \textbf{Input}: mini-batch ($X$); label set ($Y$); max\_timestep ($T$) \\
  \textbf{Output}:  updated network weights 
  \begin{algorithmic}[1]
    % \State{\textbf{begin}}
    %
    \For{$i \gets 1$ to $max\_iter$}
        \State {fetch a mini batch X}
        \For{$t \gets 1$ to $T$}  
            \State{O $\leftarrow$ PoissonGenerator(X)}
            \For{$l \gets 1$ to $L-1$} 
                \State{$(O^t_{l}, U_l^{t}) \leftarrow (\lambda, U_l^{t-1}, \textup{BNTT}_{\gamma^t}(W_{l}, O^{t-1}_{l-1}))$}
            \EndFor
            \State{\% For the final layer $L$, stack the voltage}
            \State{$U_L^{t} \hspace{-1mm} \leftarrow  \hspace{-1mm} ( U_L^{t-1},  \textup{BNTT}_{\gamma^t}(W_{l}, O^{t-1}_{L-1}))$ }
        \EndFor
        \State{\% Calculate the loss and back-propagation}
        \State{$L \leftarrow (U_L^T, Y)$}
    \EndFor
    % \State{\% Early exit}
    % \State{$t_{exit} \leftarrow (\gamma^t)$}
    % \vspace{-0.3cm}
  \end{algorithmic}
      \label{algorithm: overall}
\end{algorithm}
% \end{minipage}

% \begin{figure}
%      \centering
%          \includegraphics[width=1.0\textwidth]{ figures/gamma_fig.png}
% \caption{ Visualization of evolving $\gamma$ at conv1, conv4, conv7, and fc1 layers across all time-steps.
% Each row denotes the same layer. The experiments are conducted on VGG9 with 25 time-steps.
% }
% \label{fig:gammavisual}
% \end{figure}

\begin{figure*}
    \centering
    \includegraphics[width=0.90\textwidth]{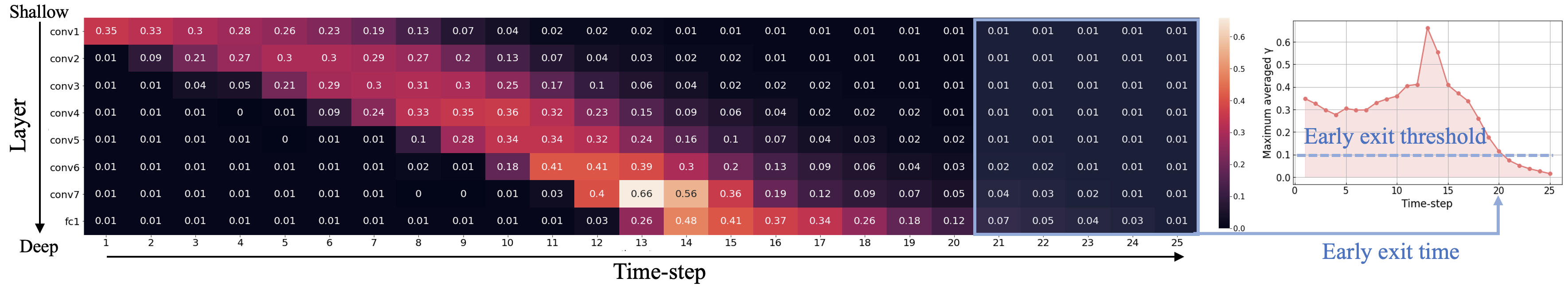}
    \vspace{-1mm}
    \caption{The average value of $\gamma^t$ at each layer over all time-steps. Early exit time can be calculated as $t=20$ since $\gamma^t$ values at every layer have lower value than threshold $0.1$ after time-step 20 (blue shaded area).
    Here, we use a VGG9 architecture on CIFAR-10.}
    \label{fig:alpha_distribution}
    \vspace{-5mm}
\end{figure*}

\subsection{Early Exit Algorithm}
\vspace{-1mm}
The main objective of early exit is to reduce the latency during inference \cite{teerapittayanon2016branchynet,panda2016conditional}.
Most previous methods \cite{wu2018spatio,lee2020enabling,sengupta2019going,rathi2020enabling,han2020rmp} accumulate output spikes till the end of the time-sequence, at inference,
since all layers generate spikes across all time-steps as shown in  Fig. \ref{fig:motivation}(a) and Fig. \ref{fig:motivation}(b).
On the other hand, learnable parameters in BNTT manipulate the spike activity of each layer to produce a peak value, which falls again (a gaussian-like trend), as shown in  Fig. \ref{fig:motivation}(c).
This phenomenon shows that SNNs using BNTT convey little information at the end of spike trains.

% by leveraging the $\gamma$ values that store the temporal dynamic information

% Unlike the existing methods in which the number of spikes converges to one value over time, learnable parameters in BNTT manipulates the spike activation of each layer to produce a peak value, and then falls again, as shown in  Fig. \ref{fig:motivation}.

% $w = \alpha exp(-\beta (t- t_{last}))$

% $L^t_{scam} = \sum_{k} W^t_k \otimes A^t_k$

Inspired by this observation, we propose a temporal early exit algorithm based on the value of $\gamma^t$.
From Eq. \ref{eq: firing_condition}, we know that a low $\gamma^t$ value increases the firing threshold, resulting in low spike activity. A high $\gamma^t$ value, in contrast, induces more spike activity.
It is worth mentioning that $(\sigma_i^t)^2$ shows similar values across all time-steps and therefore we only focus on $\gamma^t$.
Given that the intensity of spike activity is proportional to  $\gamma^t$, we can infer that spikes will hardly contribute to the classification result once $\gamma^t$ values across every layer drop to a minimum value.
Therefore, 
we measure the average of $\gamma^t$ values in each layer $l$ at every time-step, and terminate the inference when $\gamma^t$ value in every layer is below a pre-determined threshold.
{For example, as shown in Fig. \ref{fig:alpha_distribution}, we observe that all averaged $\gamma^t$ values are lower than threshold $0.1$ after $t>20$. Therefore, we define the early exit time at $t=20$.}
Note that we can determine the optimum time-step for early exit before forward propagation without any additional computation.
In summary, the temporal early exit method enables us to find the earliest time-step during inference that ensures integration  of  crucial  information, in turn reducing the inference latency without significant loss of accuracy.

\vspace{-1mm}

\subsection{Overall Optimization}
\vspace{-1mm}
Algorithm 2 summarizes the whole training process of SNNs with BNTT.
Our proposed BNTT acts as a regularizer, unlike previous methods \cite{lee2020enabling,sengupta2019going,lee2016training,rathi2020enabling} that use dropout to perform regularization.
Our training scheme is based on widely used rate coding where 
the spike generator produces a Poisson spike train (see Appendix B)  for each pixel in the image with frequency proportional to the pixel intensity \cite{roy2019towards}.
For all layers, the weighted sum of the input signal 
is passed through a BNTT layer and then is accumulated in the membrane potential.
If the membrane potential exceeds the firing threshold, the neuron generates an output spike.
For last layer, we accumulate the input voltage over all time-steps without leak, that we feed to a softmax layer to output a probability distribution. 
Then, we calculate a cross-entropy loss function and gradients for weight of each layer with the approximate gradient function.
%BN forwarding
During the training phase, a BNTT layer computes the time-dependent statistics (\ie, $\mu^t$ and $\sigma^t$) and stores the moving-average global mean and variance.
%inference
At inference, we first define the early exit time-step based on the value of $\gamma$ in BNTT. Then, the networks classify the test input (note, test data normalized with pre-computed global $\Bar{\mu}^t, \Bar{\sigma}^t$ BNTT statistics) based on the accumulated output voltage at the pre-computed early exit time-step.

\vspace{-2mm}

\section{Experiments}
\vspace{-2mm}
In this section, we carry out comprehensive experiments on public classification datasets.
Till now, training SNNs from scratch with surrogate gradient has been limited to simple datasets, \eg, CIFAR-10, due to the difficulty of direct optimization. In this paper, for the first time, we train SNNs with  surrogate gradients from scratch  and report the performance on  large-scale datasets including CIFAR-100 and Tiny-ImageNet with multi-layered network architectures. %, and ImageNet.
We first compare our BNTT with previous SNNs training methods. Then, we quantitatively and qualitatively demonstrate the effectiveness of our proposed BNTT.

\vspace{-2mm}

\subsection{Experimental Setup}
\vspace{-1mm}
We evaluate our method on three static datasets (\ie, CIFAR-10, CIFAR-100, Tiny-ImageNet) and one neuromophic dataset (\ie, DVS-CIFAR10).
\textbf{CIFAR-10} \cite{krizhevsky2009learning} consists of 60,000 images (50,000 for training / 10,000 for testing) with 10 categories. All images are RGB color images whose size are 32 $\times$ 32.
\textbf{CIFAR-100} has the same configuration as CIFAR-10, except it contains images from 100 categories.
\textbf{Tiny-ImageNet} is the modified subset of the original ImageNet dataset.
Here, there are 200 different classes of ImageNet dataset \cite{deng2009imagenet}, with 100,000 training and 10,000
validation images. The resolution of the
images is 64$\times$64 pixels.
\textbf{DVS-CIFAR10} \cite{li2017cifar10} has the same configuration as CIFAR-10. This discrete event-stream dataset is 
collected by moving the event-driven camera.
We follow the similar data pre-processing protocol and a network architecture used in previous work \cite{wu2019direct} (details in Appendix C).
% \textbf{ImageNet} (\cite{deng2009imagenet}) is  a large-scale dataset designed for practical scenarios.
% The dataset consists of labeled high-resolution 1.2 million training images and 50,000 validation images. 
Our implementation is based on Pytorch \cite{paszke2017automatic}. We train the networks with standard SGD with momentum 0.9,  weight decay 0.0005
and also apply random crop and horizontal flip to input images.
The base learning rate is set to 0.3 and we use step-wise learning rate scheduling with a decay factor 10 at 50\%, 70\%, and 90\%  of the total number of epochs. Here, we set the total number of epochs to 120, 240, 90, and 60 for CIFAR-10, CIFAR-100, Tiny-ImageNet, and DVS-CIFAR10, respectively.

\begin{table*}[t]
    \addtolength{\tabcolsep}{2.5pt}
    \centering
    \caption{Classification Accuracy (\%)  on CIFAR-10, CIFAR-100, and Tiny-ImageNet.}
    \label{table:accuracy_cifar100}
    % \vspace{-2mm}
    \resizebox{0.9\textwidth}{!}{%
    \begin{tabular}{lccccc}
        \toprule
        % \multirow{2}{30pt}{\centering Method}\:\:\:\:\:\:\:  &  \multicolumn{4}{c}{CIFAR-10} \\
        \cmidrule{2-6}
        & \:\:\:\textrm{Dataset}\:\:\: 
        & \:\:\:\textrm{Training Method}\:\:\: & \:\:\:\textrm{Architecture}\:\:\: & \:\:\:\textrm{Time-steps}\:\:\: & \:\:\:\:\:\textrm{Accuracy(\%)}\:\:\: \\
         \midrule
        Cao \textit{et al.} \cite{cao2015spiking}& CIFAR-10 & ANN-SNN Conversion & 3Conv, 2Linear & 400 & 77.4 \\
        Sengupta \textit{et al.} \cite{sengupta2019going}& CIFAR-10 & ANN-SNN Conversion & VGG16 & 2500 & 91.5  \\
        Lee \textit{et al.} \cite{lee2020enabling}& CIFAR-10& Surrogate Gradient & VGG9 & 100 & 90.4\\
        % \cite{wu2019direct} & Surrogate Gradient & 5Conv, 2Linear  & 12 & 90.53  \\
        Rathi \textit{et al.} \cite{rathi2020enabling}& CIFAR-10 & Hybrid & VGG16 & 200 & 92.0 \\ 
        Han \textit{et al.} \cite{han2020rmp}& CIFAR-10 & ANN-SNN Conversion & VGG16 & 2048 & 93.6  \\
        \textbf{\textrm{w.o. BNTT}}& CIFAR-10 & Surrogate Gradient & VGG9 & 100 & 88.7\\
        % \midrule
        \textbf{\textrm{BNTT} (\textrm{ours})}& CIFAR-10 & Surrogate Gradient & VGG9 & 25 & 90.5 \\
        \textbf{\textrm{BNTT + Early Exit} (\textrm{ours})}& CIFAR-10 & Surrogate Gradient & VGG9 & 20 & 90.3  \\
         \midrule
        %%%%%%%%%%%%%%%%%%%%%%%%%%%%%
        Sengupta \textit{et al.} \cite{sengupta2019going}& CIFAR-100 & ANN-SNN Conversion & VGG16 & 2500 & 70.9  \\
        Rathi \textit{et al.}  \cite{rathi2020enabling} & CIFAR-100 & Hybrid & VGG16 & 125 & 67.8 \\ 
         Han \textit{et al.} \cite{han2020rmp}& CIFAR-100 & ANN-SNN Conversion & VGG16 & 2048 & 70.9  \\
        \textbf{\textrm{w.o. BNTT}}& CIFAR-100 & Surrogate Gradient & VGG11 & n/a & n/a\\
        \textbf{\textrm{BNTT} (\textrm{ours})}& CIFAR-100 & Surrogate Gradient & VGG11 & 50 & 66.6\\
        \textbf{\textrm{BNTT + Early Exit} (\textrm{ours})}& CIFAR-100 & Surrogate Gradient & VGG11  & 30 & 65.8  \\
        \midrule
        %%%%%%%%%%%%%%%%%%%%%%%%%%%%%%
          Sengupta \textit{et al.} \cite{sengupta2019going}& Tiny-ImageNet & ANN-SNN Conversion & VGG11 & 2500 & 54.2\\
        % \cite{rathi2020enabling}& Tiny-ImageNet & ANN-SNN Conversion & VGG16 & ? & ?  \\
        \textbf{\textrm{w.o. BNTT}}& Tiny-ImageNet  & Surrogate Gradient & VGG11 & n/a & n/a\\
        \textbf{\textrm{BNTT} (\textrm{ours})}&  Tiny-ImageNet & Surrogate Gradient & VGG11 & 30 & 57.8 \\
        \textbf{\textrm{BNTT + Early Exit} (\textrm{ours})}&  Tiny-ImageNet & Surrogate Gradient & VGG11  & 25 & 56.8  \\
        % \midrule
        % %%%%%%%%%%%%%%%%%%%%%%%%%%%%%%%
        % \cite{orchard2015hfirst}& DVS-CIFAR10 & Random Forest & - & - & 31.0  \\
        % \cite{lagorce2016hots}& DVS-CIFAR10 & HOTS & - & - & 27.1  \\
        % \cite{sironi2018hats}&  DVS-CIFAR10 & HAT & - & - & 52.4  \\
        % \cite{sironi2018hats} &  DVS-CIFAR10 & Gabor-SNN & - & - & 24.5 \\
        % \cite{wu2019direct} &  DVS-CIFAR10 & Surrogate Gradient& - & - & 60.5 \\
        % \textbf{\textrm{w.o. BNTT}} &  DVS-CIFAR10 & Surrogate Gradient & - & - & n/a \\
        % \textbf{\textrm{BNTT} (\textrm{ours})}&  DVS-CIFAR10 & Surrogate Gradient & -  & - & 63.2  \\
        \bottomrule
        \\
    \end{tabular}%
    }
    \label{table:performance}
    \vspace{-5mm}
\end{table*}

\begin{table}[t]
\addtolength{\tabcolsep}{1.0pt}
\centering
\caption{Classification Accuracy (\%) on  DVS-CIFAR10.}
\label{table:dvs_acc}
\resizebox{0.45\textwidth}{!}{%
\begin{tabular}{lcc}
\toprule
Method &   Type  & Accuracy (\%) \\
\midrule
    Orchard \textit{et al.} \cite{orchard2015hfirst} & Random Forest & 31.0  \\
    Lagorce \textit{et al.}    \cite{lagorce2016hots} & HOTS & 27.1  \\
    Sironi \textit{et al.}    \cite{sironi2018hats} & HAT& 52.4  \\
    Sironi \textit{et al.}    \cite{sironi2018hats}  & Gabor-SNN  & 24.5 \\
    Wu \textit{et al.}    \cite{wu2019direct} & Surrogate Gradient&  60.5 \\
          \textbf{\textrm{w.o. BNTT}} & Surrogate Gradient  & n/a \\
       \textbf{\textrm{BNTT} (\textrm{ours})} & Surrogate Gradient & 63.2  \\
\bottomrule
\end{tabular}%
}
\vspace{-0.5cm}
\end{table}

\begin{figure}[t]
\begin{center}
\def\arraystretch{0.5}
\begin{tabular}{@{}c@{\hskip 0.05\linewidth}c@{}c}
\includegraphics[width=0.54  \linewidth]{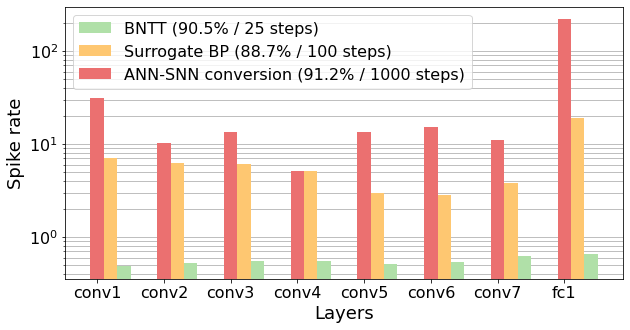} &
\includegraphics[width=0.34\linewidth]{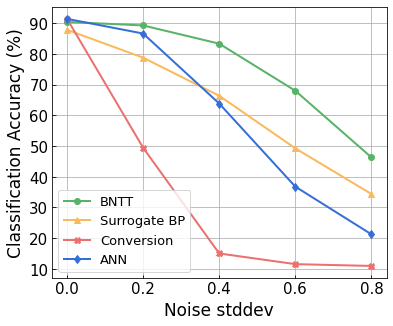} 
\\
{\hspace{2.7mm} (a) } & {\hspace{2.7mm}(b)}\\
\end{tabular}
\vspace*{-0.1in}
\end{center}
\caption{ 
(a) Visualization layer-wise spike activity (log scale) in VGG9 on CIFAR-10 dataset.
(b) Performance change with respect to the standard deviation of the Gaussian noise.
}
\vspace{-4mm}
\label{fig:energyrobust} 
\end{figure}

\subsection{Comparison with Previous Methods}

On public datasets, we compare our proposed BNTT method with previous rate-coding based SNN training methods, including
 ANN-SNN conversion \cite{han2020rmp,sengupta2019going,cao2015spiking}, surrogate gradient back-propagation \cite{lee2020enabling}, and hybrid  \cite{rathi2020enabling} methods. 
From Table \ref{table:performance}, we can observe
some advantages and disadvantages of each training method.
The ANN-SNN conversion method performs better than the surrogate gradient method across all datasets. However, they require large number of time-steps for training and testing, which is energy-inefficient and impractical in a real-time application.
The hybrid method aims to resolve this high-latency problem, but it still requires over hundreds of time-steps.
The surrogate gradient method suffers from poor optimization and hence cannot be scaled to larger datasets such as  CIFAR-100 and Tiny-ImageNet.
Our BNTT is based on the surrogate gradient method, however, it enables SNNs to achieve high performance even for more complicated datasets. 
At the same time, we dramatically reduce the latency due to the inclusion of learnable parameters and temporal statistics in the BNTT layer.
As a result, BNTT can be trained with 25 time-steps on a simple CIFAR-10 dataset, while preserving state-of-the-art accuracy.
For CIFAR-100, we achieve about $40\times$ and $2\times$ faster inference speed compared to the conversion methods and the hybrid method, respectively.
Interestingly, for Tiny-ImageNet, BNTT achieves better performance and shorter latency compared to previous conversion method.
Note that ANN with VGG11 architecture used for ANN-SNN conversion achieves 56.3\% accuracy. 
Moreover, using an early exit algorithm further reduces the latency by $\sim 20\%$, which enables the networks to be implemented with lower-latency and energy-efficiency.
It is worth mentioning that surrogate gradient method without BNTT (w.o. BNTT in Table \ref{table:performance}) only converges on CIFAR-10.
For neuromorphic DVS-CIFAR10 dataset (Table \ref{table:dvs_acc}), ANN-SNN Conversion methods are not applicable since ANNs hardly capture the temporal dynamic of a spike train. 
Using BNTT improves the stability of training compared to a surrogate gradient baseline (\ie, w.o. BNTT), and achieves state-of-the-art performance. These results show that our BNTT technique is very effective on event-driven data {and hence well-suited for neuromorphic applications}.

% CIFAR-10

% For CIFAR-100 ~ IMAGENET 
% from scratch is hard. by adding BNTT 
% dramatically reduce time-steps
% early exit - further reduce time-steps while preserving acc

% \vspace{-2mm}
\begin{figure}
     \centering
         \includegraphics[width=0.46\textwidth]{ 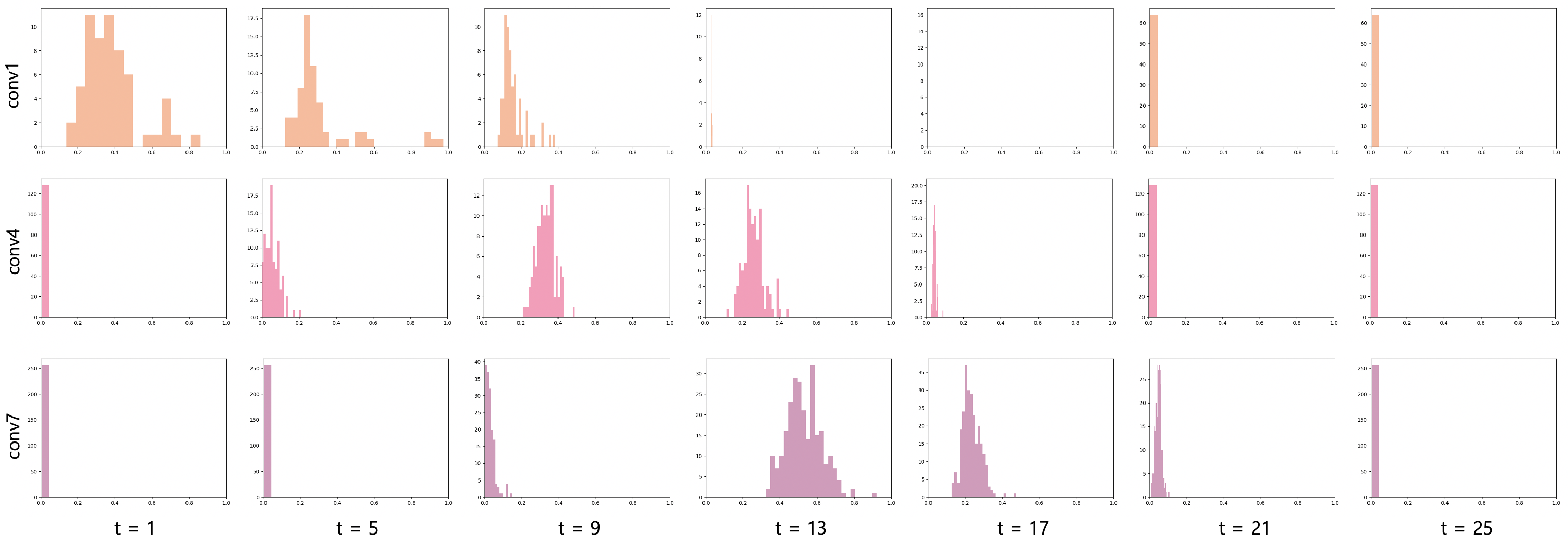}
\caption{Histogram visualization (x axis: $\gamma$ value,  y axis: frequency) at conv1 (row1), conv4 (row2), and conv7 (row3) layers in VGG9 across all time-steps.
 The experiments are conducted on CIFAR-10 with 25 time-steps.
}
\vspace{-1mm}
\label{fig:gammavisual}
\end{figure}

% \vspace{-3mm}
% \begin{table}[H]
% \addtolength{\tabcolsep}{1.5pt}
% \centering
% \caption{Energy table for 45nm CMOS process.}
% \label{table:energy_efficiency}
% % \vspace{-3mm}
% \resizebox{0.4\textwidth}{!}
% {
% \begin{tabular}{lc}
% \toprule
% Operation   & Energy(pJ)  \\
% \midrule
%     \vspace{1.3mm}
%     32bit FP MULT $(E_{MULT})$   & 3.7  \\
%     \vspace{1.3mm}
%     32bit FP ADD  $(E_{ADD})$  & 0.9  \\
%     \vspace{1.3mm}
%     32bit FP MAC $(E_{MAC})$   & 4.6 (= $E_{MULT}$ + $E_{ADD}$)  \\
%     \vspace{1mm}
%     32bit FP AC  $(E_{AC})$  & 0.9  \\
% \bottomrule
% \end{tabular}%
% }
% \vspace*{-0.1in}
% \label{table: cmos_tech}
% \end{table}
%
% \hspace{1ex}
%
\vspace{-1mm}
\begin{table}[t]
\addtolength{\tabcolsep}{1.5pt}
\centering
\caption{Energy efficiency comparison.}
\label{table:energy_efficiency}
% \vspace{-3mm}
\resizebox{0.48\textwidth}{!}{%
\begin{tabular}{lccc}
\toprule
Method &  Latency  & Accuracy (\%) & $E_{ANN}/E_{method}$ \\
\midrule
    VGG9 (ANN) &  1   & 91.5 &  1$\times$ \\
    Conversion & 1000 & 91.2  &  0.32$\times$ \\
    Conversion & 500  & 90.9 &  0.55$\times$ \\
    Conversion & 100  & 89.3 &  2.71$\times$ \\
    Surrogate Gradient&  100  & 88.7  &  1.05$\times$ \\
    \textbf{BNTT}   &  \textbf{25} & \textbf{90.5} &  \textbf{9.14$\times$} \\
\bottomrule
\end{tabular}%
}
\vspace{-0.4cm}
\label{table: energy_consumption}
\end{table}

\subsection{Energy Comparison}
\vspace{-0.2cm}
We compare the layer-wise spiking activities of our BNTT with two widely-used methods, \ie, ANN-SNN conversion method \cite{sengupta2019going} and surrogate gradient method (w.o. BNTT) \cite{neftci2019surrogate}. Note, we refer to our approach as \textit{BNTT} and standard surrogate approach w.o. BNTT as \textit{surrogate gradient} in the remainder of the text.
Specifically, we calculate the spike rate of each layer $l$, which can be defined as the total number of spikes at layer $l$ over total time-steps $T$ divided by the number of neurons in layer $l$ (see Appendix D for the equation of spike rate).
In Fig. \ref{fig:energyrobust}(a), converted SNNs show a high spike rate for every layer as they forward spike trains through a larger number of time-steps compared to other methods.
Even though the surrogate gradient method uses less number of time-steps, it still requires nearly hundreds of spikes for each layer.
Compared to these methods, we can observe that BNTT significantly improves the spike sparsity across all layers.

% P2: Quantitative comparison
More precisely, as done in previous works \cite{park2020t2fsnn,lee2016training}, we compute the  energy consumption for  SNNs in standard CMOS technology \cite{horowitz20141} as shown in Appendix D by calculating the net multiplication-and-accumulate (MAC) operations.
As the computation of SNNs are event-driven with binary \{1, 0\} spike processing, the MAC operation reduces to just a floating point (FP) addition. On the other hand, conventional ANNs still require one FP addition and one FP multiplication to conduct the same MAC operation (see Appendix D for more detail). 
Table \ref{table: energy_consumption} shows the energy efficiency of ANNs and SNNs with a VGG9 architecture \cite{simonyan2014very} on CIFAR-10.
As expected, ANN-SNN conversion yields a trade-off between  accuracy  and  energy efficiency. 
%as we  lower latency.
For the same latency, the surrogate gradient method expends higher energy compared to the conversion method. 
% This is because conversion at lower latency invokes higher firing threshold that induces sparser spike activity than the surrogate gradient.
% Compared to SNNs trained using surrogate gradients with 100 time-steps, BNTT improves the energy-efficiency significantly as we manipulate the spike activation with  learnable parameters in a BNTT layer.
It is interesting to note that even though our BNTT is trained based on the surrogate gradient method, we get $\sim 9\times$ improvement in energy efficiency compared to ANNs.
In addition, we conduct further energy comparison on  Neuromorphic architecture in Appendix E.

\vspace{-1mm}

\subsection{Analysis on Learnable Parameters in BNTT}
\label{sec: analysis_gamma}
\vspace{-1.5mm}
The key observation of our work is the change of $\gamma$ across time-steps.
To analyze the distribution of the learnable parameters in our BNTT, we visualize the histogram of $\gamma$ in conv1, conv4, and conv7 layers in VGG9 as shown in  Fig. \ref{fig:gammavisual}.
Interestingly, all layers show different temporal evolution of gamma distributions. 
For example, conv1 has high $\gamma$ values at the initial time-steps which decrease as time goes on.
On the other hand, starting from small values, the $\gamma$ values in conv4 and conv7 layers peak at $t = 9$ and $t = 13$, respectively, and then shrink to zero at later time-steps.  
% Similarly, the conv7 layer and the fc1 layer have a peak time for the maximum $\gamma$ value at $t = 12$ and $t = 14$, respectively.
Notably, the peak time is delayed as the layer goes deeper, implying that the visual information is passed through the network sequentially over a period of time similar to Fig.\ref{fig:motivation}(c).
This gaussian-like trend with rise and fall of $\gamma$ across different time-steps can support the explanation of overall low spike activity compared to other methods (Fig. \ref{fig:energyrobust}(a)).

% \textcolor{red}{ (similar to Fig. 1(c)). This gaussian-like trend with rise and fall of $\gamma$ across different time-steps.... can support the explanation of overall low spike activity (Fig. 4(a)) and in turn energy efficiency of our BNTT.}
% \textcolor{blue}{NOT UNDERSTANDABLE: As we discussed in Section \ref{sec: mathmaticalanalysis}, a large $\gamma$ value induces high spike activity, resulting in different spike activity of each layer as shown in Fig. \ref{fig:motivation}(c).
% Moreover, compared to other method that all layers generate spike across whole time-steps, each layer with BNTT does not generate spikes except for its spike time period.
% Fig. \ref{fig:energyrobust}(a) shows that layer-independent spike activity enables the networks to be implemented with a small spike rate compared to other methods.}

% As we optimize the gamma to minimize the classification loss at a training phase,  

\vspace{-1mm}

\subsection{Analysis on Early Exit}
\begin{figure}[t]
\begin{center}
\def\arraystretch{0.5}
\begin{tabular}{@{\hskip 0.015\linewidth}c@{\hskip 0.015\linewidth}c@{\hskip 0.015\linewidth}c}
\includegraphics[width=0.32  \linewidth]{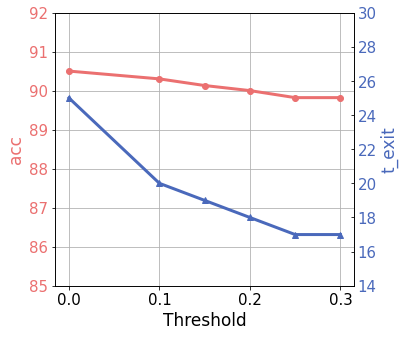} &
\includegraphics[width=0.32\linewidth]{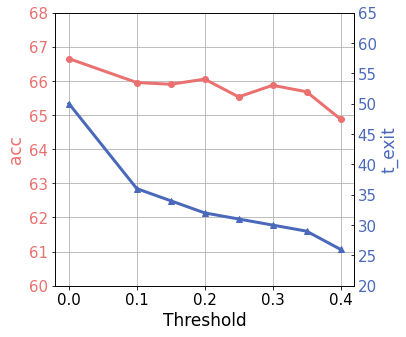} &
\includegraphics[width=0.32\linewidth]{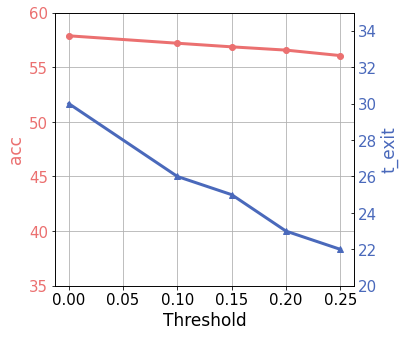} 
\\
{\hspace{2.7mm} (a) } & {\hspace{2.7mm}(b) }& {\hspace{2.7mm}(c)}\\
\end{tabular}
\vspace*{-4mm}
\end{center}
\caption{Visualization of accuracy and early exit time with respect to the threshold value for $\gamma$. (a) CIFAR-10. (b) CIFAR-100. (c) Tiny-ImageNet.}
\label{fig:early_exit}
\vspace{-5mm}
\end{figure}

\vspace{-1.5mm}
Recall that we measure the average of $\gamma$ values in each layer at every time-step, and stop the inference when all  $\gamma$  values in every layer is lower than a predetermined threshold. 
% To analyze the relationship between the threshold value and the accuracy, 
To further investigate this, we vary the predetermined threshold and show the accuracy and exit time $\textrm{T}_{exit}$ trend.
As shown in Fig. \ref{fig:early_exit}, we observe that high threshold enables the networks to infer at earlier time-steps.
Although we use less number time-steps during  inference, the accuracy drops marginally. This implies that BNTT rarely sends crucial information at the end of spike train (see Fig. \ref{fig:motivation}(c)).
Note that the temporal evolution of learnable parameter $\gamma$ with our BNTT allows us to exploit the early exit algorithm that yields a huge advantage in terms of reduced latency at inference. Such strategy has not been proposed or explored in any prior works that have mainly focused on reducing the number of time-steps during training without effectively using temporal statistics.

\vspace{-1mm}

\subsection{Analysis on Robustness}
\vspace{-1.5mm}
Finally,  we highlight the advantage of BNTT in terms of the robustness to noisy input.
% \cite{sharmin2020inherent} insisted that the SNNs inherently have more noisy resiliency compared to ANN, and the robustness of SNNs can be further enhanced by reducing the latency especially for surrogate gradient training since they reasonably capture temporal dynamics of spikes.
To investigate the effect of our BNTT for robustness, we evaluate the performance change in the SNNs as we feed in inputs with varying levels of noise. We generate the  noisy input by adding Gaussian noise $(0, \sigma^2)$ to the clean input image.
From Fig. \ref{fig:energyrobust} (b), we observe the following:
i) The accuracy of conversion method degrades considerably for $\sigma > 0.4$.
ii) Compared to ANNs, SNNs trained with surrogate gradient back-propagation shows better performance at higher noise intensity. Still, they suffer from large accuracy drops in presence of noisy inputs.
iii) BNTT achieves significantly higher performance than the other methods across all noise intensities.
This is because using BNTT decreases the overall number of time-steps which is a crucial contributing factor towards robustness \cite{sharmin2020inherent}.
These results imply that, in addition to low-latency and energy-efficiency, our BNTT method also offers improved robustness for suitably implementing SNNs in a real-world scenario.
We further analyze the robustness regarding adversarial attack \cite{goodfellow2014explaining} in Appendix F.
\vspace{-1.5mm}

\section{Conclusion}
\vspace{-2mm}
In this paper, we revisit the batch normalization technique and propose a novel mechanism for training low-latency, energy-efficient, robust, and accurate SNNs from scratch.
Our key idea is to extend the effect of batch normalization to the temporal dimension with time-specific learnable parameters and statistics.
We discover that optimizing learnable parameters $\gamma$ during the training phase enables visual information to be passed through the layers sequentially.
% This reduces the number of time-steps with increased spike sparsity while preserving accuracy.
% By transmitting visual information effectively,
% SNNs using BNTT operate at reduced number of time-steps with increased spike sparsity while preserving accuracy. We also proposed a temporal early exit algorithm by leveraging the change of $\gamma$ over different time-steps.
% Finally, through extensive experiments, we have demonstrated the efficiency and effectiveness of our BNTT on various datasets.
For the first time, we directly train SNNs on large datasets such as Tiny-ImageNet, which opens up the potential advantage of surrogate gradient-based backpropagation for future practical research in SNNs.

\appendix

\section{Appendix: Backward Gradient of BNTT}

Here, we calculate the backward gradient of a BNTT layer. 
Note that we omit the neuron index $i$ for simplicity. 
For one sample $x_b$ in a mini-batch, we compute the backward gradient of BNTT at time-step $t$: 
\begin{equation}
    \frac{\partial L}{\partial x^t_b} = \frac{\partial L}{\partial \hat{x}^t_b} \frac{\partial \hat{x}^t_b}{\partial x^t_b}
    +
    \frac{\partial L}{\partial \mu^{t}} \frac{\partial \mu^{t}}{\partial x^t_b}
    +
    \frac{\partial L}{\partial (\sigma^{t})^2} \frac{\partial (\sigma^{t})^2}{\partial x^t_b}.
    \label{append_eq: derivation}
\end{equation}
where,
\begin{equation}
    {\hat{x}^t_b}  = \frac{x^t_b - \mu^t}{\sqrt{(\sigma^{t})^2 + \epsilon}}.
\end{equation}
It is worth mentioning that we accumulate input signals at the last layer in order to remove information loss. Then we convert the accumulated voltage into probabilities by using a softmax function. Therefore, we calculate backward gradient with respect to the loss $L$, following the previous work \cite{neftci2019surrogate}.
The first term of R.H.S in Eq. (\ref{append_eq: derivation}) can be calculated as:
\begin{equation}
    \frac{\partial L}{\partial \hat{x}^t_b} \frac{\partial \hat{x}^t_b}{\partial x^t_b}
    =
    \frac{1}{\sqrt{(\sigma^{t})^2 + \epsilon}} \frac{\partial L}{\partial \hat{x}^t_b}.
    \label{eq:supply_term1}
\end{equation}
For the second term of R.H.S in Eq. (\ref{append_eq: derivation}),
\begin{equation}
    \begin{split}
     & \frac{\partial L}{\partial \mu^{t}} \frac{\partial \mu^{t}}{\partial x^t_b} 
      =
     \Big\{ \frac{\partial L}{\partial \hat{x}^t} \frac{\partial \hat{x}^t}{\partial \mu^t}
     +
     \frac{\partial L}{\partial(\sigma^{t})^2} \frac{\partial(\sigma^{t})^2}{\partial \mu^t}
     \Big\}
     \frac{\partial \mu^{t}}{\partial x^t_b} \\
     & =
     \Big\{  \sum_{j=1}^m \frac{\partial L}{\partial \hat{x}^t_j}  \frac{-1}{\sqrt{(\sigma^{t})^2 + \epsilon}}
     +
     \frac{\partial L}{\partial(\sigma^{t})^2} \frac{1}{m}
     \sum_{j=1}^{m} -2(x^t_j -\mu^{t})
     \Big\}
     \frac{\partial \mu^{t}}{\partial x^t_b} \\
      & =
     \Big\{  \sum_{j=1}^m \frac{\partial L}{\partial \hat{x}^t_j}  \frac{-1}{\sqrt{(\sigma^{t})^2 + \epsilon}}
     -2
     \frac{\partial L}{\partial(\sigma^{t})^2} (\mu^{t} - \frac{\mu^{t} m}{m})
     \Big\}
     \frac{\partial \mu^{t}}{\partial x^t_b} \\
       & =
     \Big\{  \sum_{j=1}^m \frac{\partial L}{\partial \hat{x}^t_j}  \frac{-1}{\sqrt{(\sigma^{t})^2 + \epsilon}}
     \Big\}
     \frac{\partial \mu^{t}}{\partial x^t_b} \\
        & =
     \frac{1}{m \sqrt{(\sigma^{t})^2 + \epsilon}} \Big\{ -  \sum_{j=1}^m \frac{\partial L}{\partial \hat{x}^t_j} 
     \Big\}.
     \end{split}
     \label{eq:supply_term2}
\end{equation}
For the third term of R.H.S in Eq. (\ref{append_eq: derivation}),
\begin{equation}
     \begin{split}
     & \frac{\partial L}{\partial (\sigma^{t})^2} \frac{\partial (\sigma^{t})^2}{\partial x^t_b}
      =
     \Big\{ \frac{\partial L}{\partial \hat{x}^t} \frac{\partial \hat{x}^t}{\partial (\sigma^{t})^2}
     \Big\}
     \frac{\partial (\sigma^{t})^2}{\partial x^t_b} \\
     & =
     \Big\{ -\frac{1}{2} \sum_{j=1}^m \frac{\partial L}{\partial \hat{x}^t_j}  (x^t_j - \mu^t) ((\sigma^t)^2 + \epsilon)^{-1.5}
     \Big\}
     \frac{\partial (\sigma^{t})^2}{\partial x^t_b} \\
     & =
     \Big\{ -\frac{1}{2} \sum_{j=1}^m \frac{\partial L}{\partial \hat{x}^t_j}  (x^t_j - \mu^t) ((\sigma^t)^2 + \epsilon)^{-1.5}
     \Big\}
     \frac{2(x_b -\mu^t)}{m} \\
    & =
     \Big\{ - \sum_{j=1}^m \frac{\partial L}{\partial \hat{x}^t_j}  \frac{(x^t_j - \mu^t)}{\sqrt{(\sigma^{t})^2 + \epsilon}} ((\sigma^t)^2 + \epsilon)^{-0.5}
     \Big\}
     \frac{(x_b -\mu^t)}{m \sqrt{(\sigma^{t})^2 + \epsilon}}\\
    & = 
    \Big\{ - \sum_{j=1}^m \frac{\partial L}{\partial \hat{x}^t_j}  \hat{x}^t_j ((\sigma^t)^2 + \epsilon)^{-0.5}
     \Big\}
     \frac{\hat{x}_b}{m} \\
    & = 
    \frac{\hat{x}_b}{m \sqrt{(\sigma^{t})^2 + \epsilon}}
    \Big\{ - \sum_{j=1}^m \frac{\partial L}{\partial \hat{x}^t_j}  \hat{x}^t_j 
     \Big\}.    \end{split}
     \label{eq:supply_term3}
\end{equation}

Based on Eq (\ref{eq:supply_term1}), Eq (\ref{eq:supply_term2}), and Eq (\ref{eq:supply_term3}), 
 we can reformulate Eq. (\ref{append_eq: derivation}) as:

% \begin{footnotesize}

\begin{equation}
     \begin{split}
     \scriptstyle
    & \frac{\partial L}{\partial x^t_b} = \frac{\partial L}{\partial \hat{x}^t_b} \frac{\partial \hat{x}^t_b}{\partial x^t_b}
    +
    \frac{\partial L}{\partial \mu^{t}} \frac{\partial \mu^{t}}{\partial x^t_b}
    +
    \frac{\partial L}{\partial (\sigma^{t})^2} \frac{\partial (\sigma^{t})^2}{\partial x^t_b} \\
    & = \frac{1}{\sqrt{(\sigma^{t})^2 + \epsilon}} \frac{\partial L}{\partial \hat{x}^t_b} +\frac{1}{m \sqrt{(\sigma^{t})^2 + \epsilon}} \Big\{ -  \sum_{j=1}^m \frac{\partial L}{\partial \hat{x}^t_j} 
     \Big\} \\
    & \hspace{2mm}+ \frac{\hat{x}_b}{m \sqrt{(\sigma^{t})^2 + \epsilon}}
    \Big\{ - \sum_{j=1}^m \frac{\partial L}{\partial \hat{x}^t_j}  \hat{x}^t_j 
     \Big\}  \\
    & = 
    \frac{1}{m \sqrt{(\sigma^{t})^2 + \epsilon}} 
    \Big\{  
     m \frac{\partial L}{\partial \hat{x}^t_b}
     -  \sum_{j=1}^m \frac{\partial L}{\partial \hat{x}^t_j} 
     - \hat{x}_b \sum_{j=1}^m \frac{\partial L}{\partial \hat{x}^t_j}  \hat{x}^t_j 
    \Big\}.
    \end{split}
\end{equation}
% \end{footnotesize}

To summarize, for every time-step $t$, gradients  are calculated based on the time-specific statistics of input 
signals. This allows the networks to take into account temporal dynamics for training weight connections.

\section{Appendix: Rate Coding}

\begin{figure}[h]
%   \vspace{-20pt}
  \begin{center}
    \includegraphics[width=0.45\textwidth]{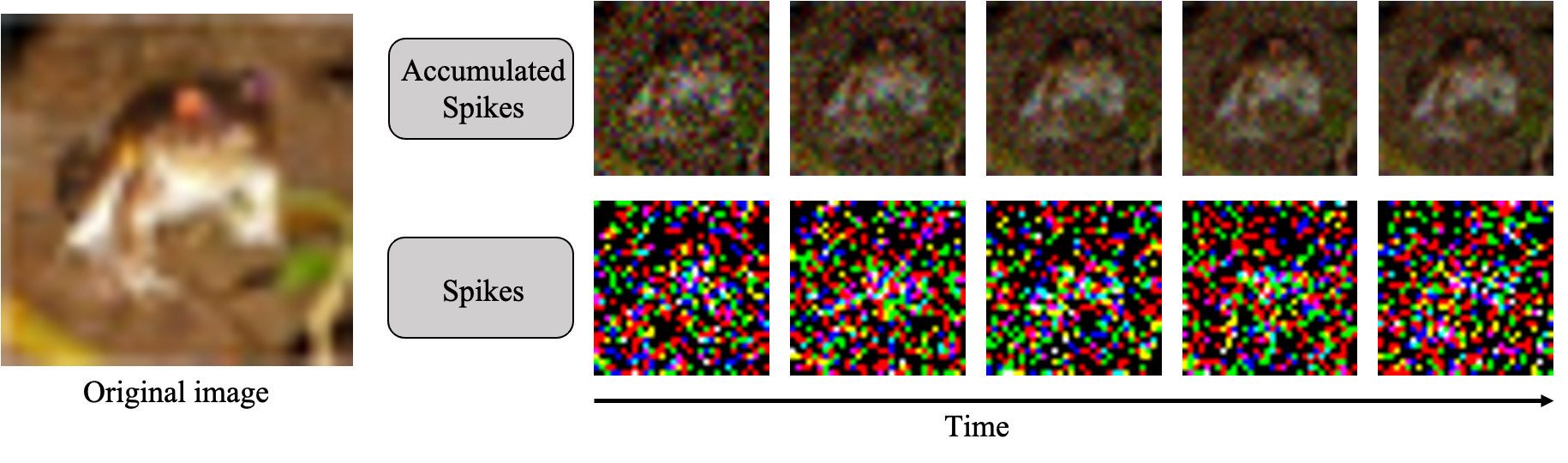}
  \end{center}
%   \vspace{-4mm}
  \caption{ Example of rate coding. As time goes on, the accumulated spikes represent similar image to original image. We use image in the CIFAR-10 dataset.
  }
%   \vspace{-2mm}
  \label{fig: possion_coding}
\end{figure}

Spiking neural networks process multiple binary spikes. 
Therefore, for training and inference, a static image needs to be converted.
There are various spike coding schemes such as rate,  temporal, and phase \cite{mostafa2017supervised,kim2018deep}.
Among them, we use rate coding due to its reliable performance across various tasks.
Rate coding provides spikes proportional to the pixel intensity of the given image.
In order to implement this, following previous work \cite{roy2019towards}, we compare each pixel value with a random number ranging between $[I_{min}, I_{max}]$ at every time-step. Here, $I_{min}, I_{max}$ correspond to the minimum and  maximum possible pixel intensity.
If the random number is greater than the pixel intensity, the Poisson spike generator outputs a spike with amplitude $1$.
Otherwise, the Poisson spike generator does not yield any spikes. 
We visualize rate coding in Fig. \ref{fig: possion_coding}. We see that the spikes generated at a given time-step is random. However, as time goes on, the accumulated spikes represent a similar result to the original image.

\section{Appendix: DVS-CIFAR10 dataset}

\begin{figure}[h]
%   \vspace{-20pt}
  \begin{center}
    \includegraphics[width=0.2\textwidth]{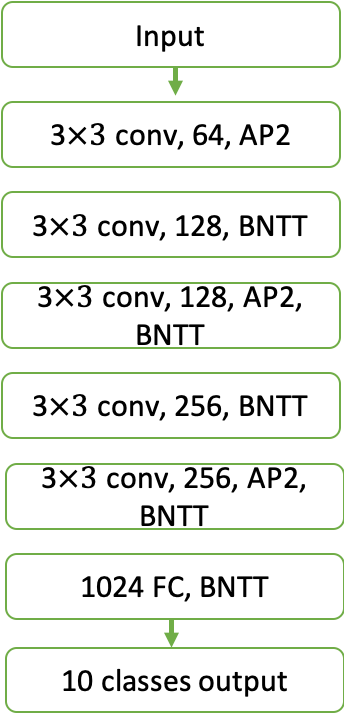}
  \end{center}
%   \vspace{-4mm}
  \caption{ Illustration of network structures for DVS dataset. Here, AP denotes average pooling, FC denotes fully connected configuration.}
%   \vspace{-2mm}
  \label{fig:dvs_architecture}
\end{figure}

 On DVS-CIFAR10, following \cite{wu2019direct}, we downsample the size of the  128 ×128 images to 42×42.
Also, we divide the total number of time-steps available from the original time-frame data into 20 intervals and accumulate the spikes  within each interval.
We use a similar architecture as previous work \cite{wu2019direct}, which consists of a  5-layerered feature extractor and a classifier.
The detailed architecture is shown in Fig. \ref{fig:dvs_architecture} in this appendix.

\section{Appendix: Energy Calculation}

In this appendix section, we provide the details of energy calculation discussed in Section 4.3 in the main paper.
The total computational cost is proportional to  the total number of floating point operations (FLOPS).
This is approximately the same as the number of Matrix-Vector Multiplication (MVM) operations.
For layer $l$ in ANNs, we can calculate FLOPS as:
\begin{equation}
FLOPS_{ANN}(l) = \left\{\begin{matrix} &k^2 \times O^2 \times C_{in} \times C_{out},  \hspace{4.5mm}  \textup{if $l$: Conv},
\\ 
&\hspace{1mm}C_{in} \times C_{out}, \hspace{21mm} \textup{if $l$: Linear}.
\end{matrix}\right.
\end{equation}
Here, $k$ is kernel size. $O$ is output feature map size. 
$C_{in}$ and $C_{out}$ are input and output channels, respectively.
For SNNs, we first define spiking rate $R_s(l)$ at layer $l$ which is the average firing rate per neuron.
\begin{equation}
    R_s(l) = \frac{\textup{$\#$spikes of layer $l$ over all timesteps}}{\textup{ $\#$neurons of layer $l$}}.
\end{equation}
Since neurons in SNNs nly consume energy whenever the neurons spike, we multiply the spiking rate $R_s(l)$  with FLOPS to obtain the SNN FLOP count. 
\begin{equation}
    FLOPS_{SNN}(l) = FLOPS_{ANN}(l) \times R_s(l).
\end{equation}
Finally, total inference energy of ANNs ($E_{ANN}$) and SNNs ($E_{SNN}$) across all layer can be obtained.
\begin{equation}
    E_{ANN} = \sum_{l} FLOPS_{ANN}(l)  \times E_{MAC}.
\end{equation}
\begin{equation}
    E_{SNN} = \sum_{l} FLOPS_{SNN}(l) \times E_{AC}.
\end{equation}

The $E_{AC}$, $E_{MAC}$ values are calculated using a standard 45 nm CMOS process \cite{horowitz20141} as shown in Table 1.

\begin{table}[h]
    \addtolength{\tabcolsep}{1.5pt}
    \centering
    \caption{Energy table for 45nm CMOS process.}
    % \vspace{-3mm}
    \vspace{1mm}
    \resizebox{0.4\textwidth}{!}{%
    \begin{tabular}{lc}
    \toprule
    Operation   & Energy(pJ)  \\
    \midrule
        32bit FP MULT ($E_{MULT}$)   & 3.7  \\
        32bit FP ADD  ($E_{ADD}$)  & 0.9  \\
        32bit FP MAC ($E_{MAC}$)   & 4.6 (= $E_{MULT} + E_{ADD}$)  \\
        32bit FP AC  ($E_{AC}$)  & 0.9  \\
    \bottomrule
    \end{tabular}
    }
    % \vspace*{-0.1in}
    \label{table: cmos_tech}
    \end{table}

\section{Appendix: Energy Comparison in Neuromorphic Architecture}

\begin{table}[h]
    \addtolength{\tabcolsep}{1.5pt}
    \centering
    \caption{Normalized energy comparison on neuromorphic architecture:  TrueNorth\cite{akopyan2015truenorth}.
    We set conversion as a reference for normalized energy comparison. We conduct experiments on CIFAR-10 with a VGG9 architecture.
    }
    \vspace{1mm}
    \resizebox{0.45\textwidth}{!}{%
    \begin{tabular}{lcccc}
    \toprule
    Method   & Time-steps  & \#Spikes ($10^4$) & Energy \cite{akopyan2015truenorth}   \\
    \midrule
        Conversion  & 1000 & 419.30 &  1\\
        Surrogate & 100 & 141.96 & 0.3384\\
        BNTT   & 25& 13.106 &  0.0312\\
    \bottomrule
    \end{tabular}%
    }
    \vspace*{-0.1in}
    \label{table: neuromorphic_energy}
    \end{table}
    
We further show the energy-efficiency of BNTT in a neuromorphic architecture, TrueNorth \cite{akopyan2015truenorth}.
Following the previous work \cite{park2020t2fsnn,moradi2018impact}, we compute the normalized energy, which can be classified into dynamic energy ($E_{dyn}$) and static energy ($E_{sta}$).
The $E_{dyn}$ value corresponds to the computing cores and routers, and $E_{sta}$ is for maintaining the state of the CMOS circuit.
The total energy consumption can be calculated as   \#Spikes $\times$ $E_{dyn}$ + \#Time-step $\times$ $E_{sta}$, where  ($E_{dyn}$, $E_{sta}$) are (0.4, 0.6).
In Table \ref{table: neuromorphic_energy}, we show that our BNTT has a huge advantage in terms of energy efficiency in neuromorphic hardware.

\section{Appendix: Adversarial Robustness}

\begin{figure}[h]
%   \vspace{-20pt}
  \begin{center}
    \includegraphics[width=0.4\textwidth]{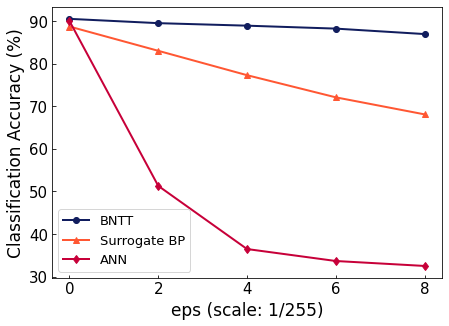}
  \end{center}
%   \vspace{-4mm}
  \caption{ Classification accuracy with respect to the intensity of FGSM attack (eps). }
%   \vspace{-2mm}
  \label{fig:FGSM attack}
\end{figure}

In order to further validate the robustness of BNTT, we conduct experiments on adversarial inputs. 
We use FGSM \cite{goodfellow2014explaining} to generate adversarial samples for ANN. 
For a given image $x$, we compute the loss function $\mathcal{L}(x,y)$ with the ground truth label $y$. The objective of FGSM attack is to change the pixel intensity of the input image that maximizes the cost function:
\begin{equation}
    x_{adv} = x + \epsilon \times sign(\nabla_{x}  \mathcal{L}(x,y)).
\end{equation}
We call $x_{adv}$ as ``adversarial sample". Here, $\epsilon$ denotes the strength of the attack.
To conduct the FGSM attack for SNN, we use the SNN-crafted FGSM method proposed in \cite{sharmin2020inherent}.
In Fig. \ref{fig:FGSM attack}, we show the classification performance for varying intensities of FGSM attack.
The SNN approaches (\eg, BNTT and Surrogate BP) show more robustness than ANN due to the temporal dynamics and stochastic neuronal functionality.
We highlight that our proposed BNTT shows much higher robustness compared to others.
Thus, we assert that BNTT improves robustness of SNNs in addition to energy efficiency and latency.

\section{Appendix: Comparison with Layer Norm}

Layer Normalization (LN) \cite{ba2016layer} proposed  the optimization method for recurrent neural networks (RNNs).
They asserted that directly applying BN layers is hardly applicable since RNNs vary with the length of the input sequence.
To this end, a LN layer calculates the mean and variance for each single layer.
As SNNs also take the time-sequence data as an input, we compare our BNTT wit Layer Normalization in Table \ref{table:layernorm}.
For all experiments, we use a VGG9 architecture. Also,  we set a base learning rate to 0.3 and we use step-wise learning rate scheduling as described in Section 4.1 of our main manuscript. The results show that BNTT is more suitable structure to capture the temporal dynamics of  
Poisson encoding spikes.

\begin{table}[t]
    \addtolength{\tabcolsep}{1.5pt}
    \centering
    \caption{Comparison with Layer Normalization on CIFAR-10 dataset.}
    \label{table:layernorm}
    \vspace{1mm}
    \resizebox{0.32\textwidth}{!}{%
    \begin{tabular}{lc}
    \toprule
    Method   & Acc (\%) \\
    \midrule
        % \vspace{1.3mm}
        Layer Normalization \cite{ba2016layer}   & 75.4  \\
        % \vspace{1.3mm}
        BNTT  & 90.5  \\
    \bottomrule
    \end{tabular}%
    }
    % \vspace*{-0.1in}
\end{table}

{\small
\bibliographystyle{ieee_fullname}
\bibliography{egbib}
}

\end{document}